# Machine Learning for Postprocessing Ensemble Streamflow Forecasts

Sanjib Sharma[1*], Ganesh Raj Ghimire[2], and Ridwan Siddique[3]

[1]Earth and Environmental Systems Institute, The Pennsylvania State University, University Park, PA 16801, USA (svs6308@psu.edu)

[2]Environmental Sciences Division, Oak Ridge National Laboratory, Oak Ridge, TN 37831, USA (ghimiregr@ornl.gov)

[3]Energy Systems and Climate Analysis, Electric Power Research Institute (EPRI), Washington D.C. 20036, USA (ridwan.siddique@gmail.com)

[*]Corresponding author: Sanjib Sharma (svs6308@psu.edu)

## Abstract

Skillful streamflow forecasts can inform decisions in various areas of water policy and management. We integrate numerical weather prediction ensembles, distributed hydrological model and machine learning to generate ensemble streamflow forecasts at medium-range lead times (1 - 7 days). We demonstrate a case study for machine learning application in postprocessing ensemble streamflow forecasts in the Upper Susquehanna River basin in eastern United States. Our results show that the machine learning postprocessor can improve streamflow forecasts relative to low complexity forecasts (e.g., climatological and temporal persistence) as well as standalone hydrometeorological modeling and neural network. The relative gain in forecast skill from postprocessor is generally higher at medium-range timescales compared to shorter lead times; high flows compared to low-moderate flows, and warm-season compared to the cool ones. Overall, our results highlight the benefits of machine learning in many aspects for improving both the skill and reliability of streamflow forecasts.

## 1. Introduction

Reliable and skillful streamflow forecasts are crucial for informing decisions related to water resources management, water supply planning, and preparedness against extreme events. Ensemble prediction system is becoming increasingly popular for streamflow forecasting (Cloke and Pappenberger, 2009; Demargne et al. 2014; Troin et al. 2021; Liu et al. 2022; Hapuarachchi et al. 2022), as they have demonstrated substantial improvements over the single-valued deterministic forecasts (Siddique and Mejia, 2017). An ensemble prediction system can provide multiple realizations of possible streamflow conditions enabling decision-makers to have a better idea of the likelihood of a specific future event (e.g., the probability of exceeding a flood threshold). More specifically, ensembles can provide an estimate of predictive uncertainty that can help decision-makers to determine the level of confidence they can place in the forecast.

Within an ensemble prediction system, hydrological models are generally forced with ensemble meteorological forecasts from weather prediction models (National Weather Service, 2022; Alfieri et al. 2014; Pagano et al., 2016; Demargne et al. 2014; Siddique and Mejia, 2017 ; Zhang et al. 2020; Hapuarachchi et al. 2022) as opposed to streamflow simulations, which are often generated by hydrological models forced with meteorological observations (Konapala et al. 2020; Kratzert et al. 2018; Feng et al. 2020). In the United States, the NOAA's National Weather Service River Forecast Centers are implementing the Hydrological Ensemble Forecast Service (HEFS) to incorporate meteorological ensembles into their flood forecasting operations (Brown et al. 2014; Kim et al. 2018; National Weather Service, 2022). A few other examples include the European Flood Awareness System from the European Commission (Alfieri et al. 2014) and the Flood Forecasting and Warming Service from the Australia Bureau of Meteorology (Pagano et al. 2016) which have adopted the ensemble paradigm. However, the shortcomings in hydrologic model structure and parameters, inadequate representation of physical processes, and biased meteorological forcing can introduce biases in ensemble streamflow forecasts (Brown et al., 2014). Uncertainties in the ensemble prediction system are of both meteorological and hydrological origins (Demargne et al. 2014); hence, the bias structure can be different in an ensemble prediction system than in hydrologic simulations (Siddique and Mejia, 2017). Correcting these residual errors and biases can improve the skill and reliability of streamflow forecasts (Regonda et al. 2013).

Hydrologic postprocessors are used to quantify total predictive uncertainty and correct forecast biases (Alizadeh et al. 2020; López López et al. 2014; Regonda et al. 2013; Seo et al.

2006). The ensemble postprocessor (EnsPost) is an integral part of the NOAA's HEFS system (Seo et al. 2006). EnsPost is a typical Hydrologic Model Output Statistics (HMOS) approach that relies on the combination of probability matching and autoregressive modeling to correct streamflow forecast biases. Several other HMOS postprocessors have been proposed, including Logistic regression (Duan et al. 2007), Quantile regression (Koenker, 2005), Autoregressive exogenous model (Regonda et al. 2013), and General linear model (Zhao et al. 2011). However, HMOS postprocessors (i) show limited performance across longer lead times (beyond ~day 3), particularly for random errors (Sharma et al. 2019); (ii) are often irrelevant for hydrologic conditions (e.g., extreme events) outside the training period (Siddique and Mejia, 2017); and (iii) fail to capture the nonlinear dynamics in hydrometeorological predictions (Regonda et al. 2013).

Another avenue to address shortcomings in postprocessing is machine learning. Machine learning algorithms identify the nonlinear patterns in a historical dataset during training and use those patterns to correct for systematic ensemble biases. Postprocessing the highly nonlinear Numerical Weather Prediction (NWP) outputs is one of many applications of machine learning techniques (Grönquist et al. 2021; Kirkwood et al. 2021; Loken et al. 2019, 2020; Rasp and Lerch, 2018). Machine learning applications in hydrology broadly span rainfall-runoff modeling (Van et al. 2020), groundwater modeling (Wunsch et al. 2021), hydrologic predictions (Panahi et al. 2022), and climate change impact assessment (Bai et al. 2021), among others. For instance, Kratzert et al. (2019) demonstrated the ability of the Long Short-Term Memory (LSTM) neural network in simulating streamflow at ungauged basins based on static catchment characteristics. Konapala et al. (2020) built a hybrid framework by coupling neural network with the hydrological model output to postprocess streamflow simulations in diverse catchments across the conterminous United States. Recently, Frame et al. (2021) demonstrated the application of LSTM networks in postprocessing National Water Model output. Although recent studies (Konapala et al. 2020; Kratzert et al. 2018; Feng et al. 2020; Tyralis et al. 2019; Xiang and Demir, 2020; Sikorska-Senoner and Quilty, 2021; Frame et al. 2021; Cho and Kim, 2022) have shown important applications of machine learning to improve various aspects of hydrologic modeling and simulations, their ability to improve the skill and reliability of streamflow forecasts obtained from the ensemble prediction system has not been examined rigorously (Lee and Ahn, 2021; Alizadeh et al. 2021).

Machine learning configurations in forecasting mode range from standalone (Cheng et al.2020) to hybrid (Hunt et al. 2022) to postprocessing (Liu et al., 2021). Chang et al., (2020) used an artificial neural network and LSTM to forecast streamflow at daily and monthly scales. Hunt et al. (2022) developed a hybrid LSTM configuration trained with catchment-mean meteorological and hydrological variables to produce skillful medium-range streamflow forecasts across various climate regions over the western United States. Liu et al. (2021) integrated meteorological forecasts, hydrological modeling and machine learning to improve flood forecasting over a cascade reservoir catchment. Overall, studies indicate that the relative effects of machine learning depend strongly on the forecasting system (e.g., forcing, hydrological model), forecasting conditions (e.g., lead time, study area, flow threshold, season), and machine learning configurations (e.g., standalone, hybrid, postprocessor) underscoring the research need for continuous rigorous verification of new forecasting systems that incorporate numerical weather prediction, hydrological modeling and machine learning.

We demonstrate the application of a machine learning-based postprocessing approach for medium-range ensemble streamflow forecasts generated using a hydrologic ensemble prediction system. We use the National Centers for Environmental Prediction Global Ensemble Forecast System Reforecast version 2 (GEFSRv2; Hamill et al. 2013) to force a spatially distributed hydrological model and generate raw ensemble streamflow forecasts at medium-range lead times (1 - 7 days). The raw ensemble streamflow forecast is used to configure the neural network postprocessor, which we describe in detail in Section 2. Then, we assess the quality of machine learning postprocessed forecasts relative to the climatological, persistence, deterministic, and raw ensemble forecasts. We also compare the performance of machine learning based postprocessor with the standalone configurations of hydrometeorological modeling and machine learning. Here we address two main questions: i) How skillful are the machine learning postprocessed ensemble streamflow forecasts at medium-range forecast lead times? and What forecast conditions (e.g., lead time, season, and flow threshold) benefit the most from machine learning?

## 2. Methods and Materials

### 2.1. Hydrologic Modeling

We demonstrate a case study for postprocessing ensemble streamflow forecasts in the Upper Susquehanna River basin in the eastern United States (Figure 1). We select the US Geological Survey (USGS) gage station 01510000 as a forecast location where frequent and severe floods are a major concern (Gitro et al. 2014). The selected station is located in the Ostelic River at Cincinnatus representing a drainage area of ~381 km². The climate in the Upper Susquehanna River basin is relatively humid, with snowy cold winters and warm summers. Flooding during the cool season is often associated with a positive North Atlantic Oscillation phase that can result in generally increased precipitation amounts accompanied by snowmelt runoff (Durkee et al. 2008). During the summer, convective precipitation leads to greater variation in streamflow. However, more recent high-water levels were associated with different flood-generating mechanisms (Smith et al. 2010), including late winter-early spring extratropical systems (April 2005), warm-season convective systems (June 2006), and tropical storm Lee (September 2011).

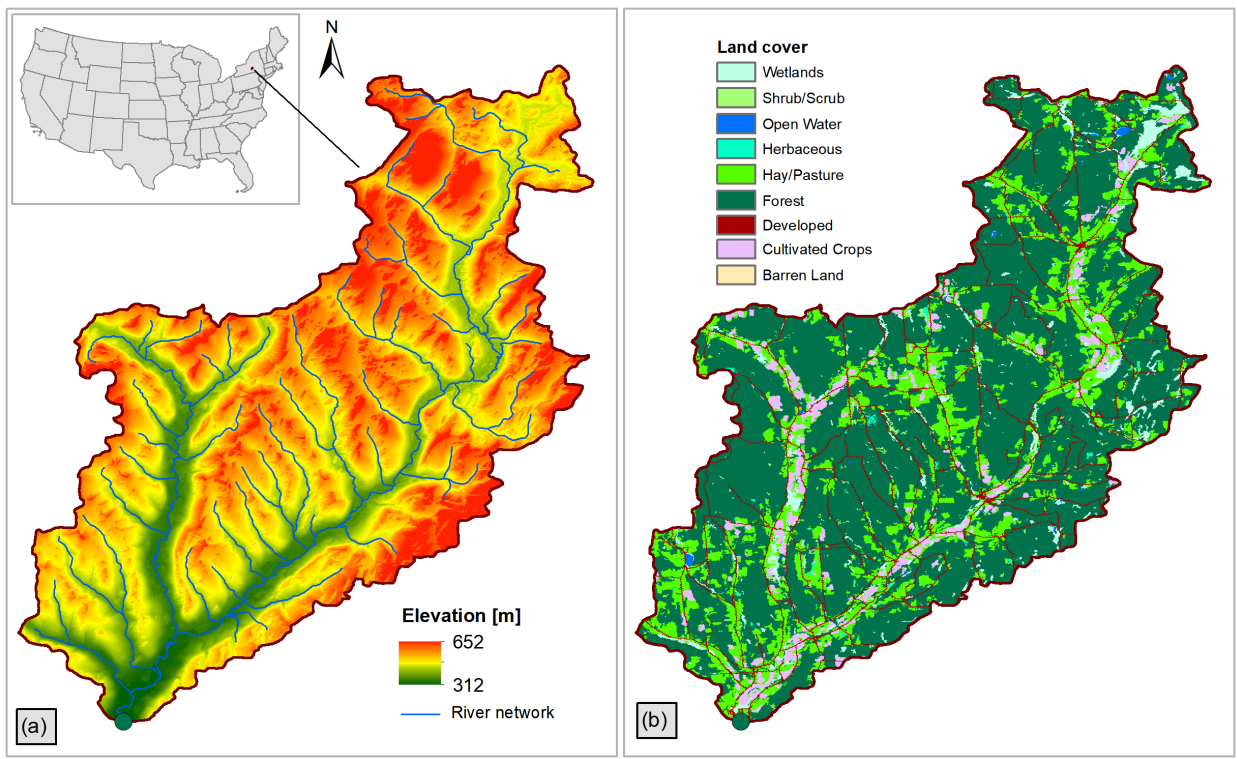

Figure 1: Map of the study area showing a) topography, stream network, location of the selected gauges station (green dot), and b) land cover types. The iset map shows the approximate location of the study area in the United States.

We employ a Regional Hydrological Ensemble Prediction System (RHEPS; Sharma et al. 2019) to generate raw ensemble streamflow forecasts at the selected USGS gage station (i.e.,

USGS 01510000). The RHEPS is an ensemble-based river forecasting system aimed at enhancing hydrologic forecasting at a regional spatial scale by integrating new system components within a verifiable scientific and experimental setting. RHEPS uses multisensor precipitation estimates and gridded near-surface air temperature observation datasets (Siddique and Mejia, 2017) to run the hydrological model in simulation mode and to train the standalone LSTM neural network. RHEPS uses precipitation and near-surface temperature ensemble forecasts from the National Centers for Environmental Prediction Global Ensemble Forecast System Reforecast version 2 (GEFSRv2) (Hamill et al. 2013) as forcing to the NOAA's Hydrology Laboratory Research Distributed Hydrologic Model (Koren et al. 2004) and generates ensemble streamflow forecasts. The GEFSRv2 is an 11-member ensemble forecast generated by perturbing the NWP model initial conditions using the ensemble transform technique with rescaling. Hydrologic model runs were initiated once a day at 00:00 Coordinated Universal Time. Each forecast cycle consists of 6-hourly streamflow forecasts that extend from day 1 to day 7. We run a hydrologic model in a fully distributed manner at a spatial resolution of 2 km × 2 km. Streamflow forecast consists of 11 ensemble members, one of which is an unperturbed member, and the rest are members generated with GEFSRv2 perturbed initial conditions. Deterministic forecasts are obtained by forcing a hydrologic model with unperturbed GEFSRv2 members. Streamflow forecasts from the RHEPS are produced for medium-range time scales and the period of 2004 - 2012. Daily streamflow observations for the selected location were obtained from the USGS. The streamflow observations are used to verify the raw and LSTM postprocessed ensemble streamflow forecasts.

## 2.2. LSTM Network

The LSTM (Hochreiter and Schmidhuber, 1997) is a special type of recurrent neural network that can learn the long-term nonlinear temporal dependencies between variables through the combination of unique structures, cells, and gates (Figure 2). The LSTM is better equipped for multi-step forecasts than some conventional machine learning architectures without cell memory (Konapala et al. 2020; Kratzert et al. 2018). Deep learning-based LSTM can provide sizeable computational benefits by using parallel processing through a graphical processing unit and advanced optimization algorithms. Previous studies have demonstrated that the LSTM deep learning model more often produces hydrologic simulations with better prediction skills than process-based models (Kratzert et al. 2018) as well as other Machine learning models such as multilayer perception, artificial neural networks, and support vector machine (Zhang et al 2018, Kumar et al. 2019; Rahimzad et al. 2021). Rainfall-runoff transformation is a scale-dependent process i.e., across both space and time. Predicting streamflow, therefore, often requires input from longer timescale processes such as groundwater storage, snowmelt, and soil moisture, among others. Unlike traditional recurrent neural networks, the ability of LSTM to learn long-term temporal dependencies between variables and simulate such long-term memory effects makes it particularly suitable for rainfall-runoff modeling (Konapala et al. 2020; Kratzert et al. 2018; Shen, 2018). The LSTM can be adapted to both catchment attributes and meteorological forcing making it appropriate for simulating rainfall-runoff dynamics across basins (Konapala et al. 2020; Kratzert et al. 2018). Here, we integrate LSTM with a distributed hydrologic model in a hybrid framework to account for the key hydrologic processes that a standalone LSTM sometimes misses (Konapala et al. 2020; Kratzert et al. 2018). We also compare the prediction skill of the LSTM postprocessor relative to standalone hydrometeorological modeling and LSTM configuration (Figure 2).

The LSTM network consists of a set of recurrently connected memory blocks containing a cell state, forget gate, input gate, output gate, and hidden state (Figure 2). Let, the input sequence be $X = [x_1, \ldots, x_t]$, where $x_t$ refers to a vector containing input features at any time step $t$. The first gate is the forget gate ($f_t$), which determines how much the former state is retained in the cell:

$$f_t = \sigma \ (W_f x_t \ + \ U_f \ h_{t-1} + b_f), \tag{1}$$

where $\sigma$ is the logistic sigmoid function, $W_f$ represents the weight connecting forget gate with the input, $U_f$ denotes the weights from the forget gate to the hidden state, $h_{t-1}$ refers to the hidden state at time step $t-1$, and $b$ is a bias vector. A potential update vector for the cell state is computed from the current input and the last hidden state:

$$\widetilde{g_t} = tanh \ (W_g x_t \ + \ U_g \ h_{t-1} + b_g), \tag{2}$$

where $\widetilde{g_t}$ is the cell input, and $tanh$ is the hyperbolic tangent function.

The second gate is the input gate ($i_t$), which controls the extent to which a new input flows into the cell:

$$i_t = \sigma \ (W_i x_t \ + \ U_i \ h_{t-1} + b_i), \tag{3}$$

where $W_i$ represents the weight connecting the input gate with the input; and $U_i$ denotes the weight from the input gate to the hidden state.

The cell state ($c_t$) characterizes the memory of the system, and is updated using:

$$c_t = f_t \ \Phi \ c_{t-1} + \ i_t \ \Phi \ \widetilde{g_t} \ , \tag{4}$$

where $\Phi$ represents element-wise multiplication.

Finally, the output gate $(o_t)$ controls the information of the cell state that flows into the new hidden state:

$$o_t = \sigma\,(W_o x_t\,+\,U_o h_{t-1}+b_o),\qquad\qquad(5)$$

where $W_o$ represents the weights connecting the output gate with the input, and $U_o$ denotes the weight from the output gate to the hidden state.

Together with Equations (4) and (5), a new hidden state $(h_t)$ is computed as:

$$h_t =\, o_t\,\Phi tanh\ (c_t).\qquad\qquad(6)$$

The intuition behind this network is that the cell states behave as the memory unit to remember useful information through the different operations of each gate.

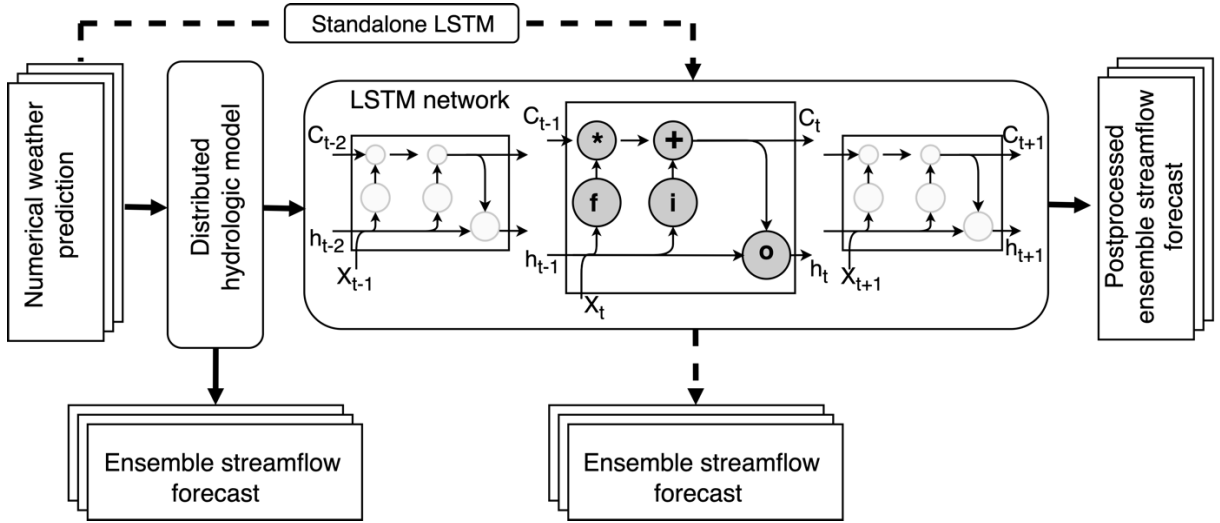

**Figure 2:** LSTM-enabled ensemble streamflow forecasting framework. The framework consists of numerical weather prediction output to force the distributed hydrologic model and produce raw ensemble streamflow forecasts. Raw ensemble streamflow forecast is used as an input to the LSTM model. We also show here the internals of the LSTM cell, where *f* stands for the forget gate, *i* for the input gate, *o* for the output gate, $c_t$ denotes the cell state at time step *t,* and $h_t$ denotes the hidden state.

We trained and tested two different configurations of the LSTM neural network to generate daily ensemble streamflow forecasts. First, a standalone LSTM neural network is trained with meteorological observations, including multisensor precipitation estimates and gridded surface temperature. The trained standalone LSTM is forced by the precipitation and near surface temperature ensemble forecasts from the GEFSRv2 to produce ensemble streamflow forecasts. Second, we trained and tested the LSTM model that simulates streamflow forecast residuals based on the time series of raw ensemble streamflow forecasts generated by forcing a distributed hydrologic model with GEFSRv2. Raw ensemble streamflow forecast residual is the difference between raw ensemble member forecasts and observations. Here we implemented LSTM separately for each lead time (day 1 to 7) and each ensemble member. The estimated residual is then added to the corresponding raw ensembles from the hydrologic model to generate the postprocessed ensemble streamflow forecasts. We implemented LSTM for the period of 2004-2012, with 6 years (2004-2009) for training the LSTM network and the remaining 3 years (2010-2012) for validation. For this, we generated 6-hourly streamflow forecasts since this is a temporal resolution often used in medium-range operational forecasting in the U.S. Note that we computed the mean daily flow forecast from the 6-hourly flow forecasts. The LSTM is then configured to the mean daily forecast residuals for forecast lead times from day 1 to 7.

The LSTM model performance is influenced by the choice of hyperparameters, including optimization algorithm, number of hidden layers, number of samples propagated through the network for each gradient update (batch size), and number of trained epochs. For efficient learning and faster convergence, all input and output features are first normalized to the range [0,1] without altering the shape of the original distribution. We considered a single-layer network with 20 hidden states. We select a batch size of 32 (Bengio, 2012), and mean-squared error as a loss function. Adam algorithm (Kingma and Ba, 2014) is applied to optimize and update the network weights. We trained the network for 30 epochs. These values are determined through several sensitivity test runs for different lead times and ensemble members. These selections are carefully done to minimize the loss function, prevent overfitting issues, and allow the network to have robust learning.

### 2.3. Forecast verification

We verified streamflow forecasts against observed streamflow at the basin outlet (USGS 01510000). Verification was performed conditionally upon a lead time, flow threshold, and seasonality. We employed Nash-Sutcliffe Efficiency (NSE), Root-Mean-Square Error (RMSE), and Percent bias (Pbias) for assessing the quality of ensemble mean forecasts. The NSE (Nash and Sutcliffe, 1970) is the ratio of the residual variance to the initial variance. The range of NSE can vary between negative infinity to 1, with 1 representing the optimal value and values should be larger than 0 to indicate minimally acceptable performance. RMSE measures the square root of the mean of the squared errors. Pbias measures the average tendency of the model predictions to be larger or smaller than their observed counterparts. We benchmarked the performance of raw and LSTM postprocessed ensemble mean forecasts relative to low-complexity forecasting approaches such as climatology and persistence-based forecasts (Ghimire and Krajewski, 2020; Krajewski et al. 2020). The persistence-based forecast implies that the streamflow behavior does not change over the forecast lead time. The approach is tied to the concept of "memory" of the system. Here we explore two approaches: simple persistence and anomaly persistence in streamflow forecasting. Simple persistence assumes that the streamflow forecast at one step ahead is dependent on the streamflow at the current time step. The anomaly persistence forecast scheme assumes that streamflow anomalies persist over the lead time. Note that we computed the anomaly with reference to the climatological average.

To investigate the probabilistic attributes of forecasts, we examined the Brier Skill Score (BSS; Brier, 1950) and reliability diagram (Murphy, 1973) of ensemble streamflow forecasts. A BSS is closely related to the Brier Score (BS), which measures the square error of discrete probability forecast and a corresponding observed outcome [0, 1], averaged over all pairs of forecasts and observations. The BS can be expressed as follows:

$$BS = \frac{1}{n} \sum_{i=1}^{n} \left[ F_{f_i}(z) - F_{Y_i}(z) \right]^2, \tag{7}$$

where the probability of $f_i$ to exceed a fixed threshold $z$ is given by,

$$F_{f_i}(z) = P_r[f_i > z], \tag{8}$$

$n$ is again the total number of forecast-observation pairs and

$$F_{f_i}(z) = \begin{cases} 1, & Y_i > z \\ 0, & otherwise. \end{cases} \tag{9}$$

The ideal value of BS is 0. To compare the skill score of the main forecast system with respect to the reference forecast, it is convenient to define the Brier Skill Score (BSS) as:

$$BSS = 1 - \frac{BS_{main}}{BS_{reference}}, \tag{10}$$

where $BS_{main}$ and $BS_{reference}$ are the BS values for the main forecast system (i.e., the system to be evaluated) and the reference forecast system, respectively. We compute the BSS with reference to sampled climatology of historical observed flow. Any positive values of the BSS [0, 1] indicate that the main forecasting system performed better than the reference forecasting system. A BSS of 0 indicates no skill (i.e., the same skill as the reference system) and a BSS of 1 indicates perfect skill.

The reliability diagram plots the average observed probability of occurrence of an event given the forecast probability, against its forecast probability of occurrence (Jolliffe and Stephenson, 2012; Wilks, 2011). The reliability diagram shows the full joint distribution of forecasts and observations for a discrete event and conveys information about the quality of forecasts in different ranges of the forecast probability distribution. For the forecast values portioned into bin $B_k$ and defined by the exceedance of threshold $z$, the average forecast probability can be expressed as:

$$\bar{F}_{f_k}(z) = \frac{1}{|I_k|} \sum_{I_k} F_{f_i}(z), \text{ where } I_k = \{i: f_i \in B_k\}, \tag{11}$$

where $I_k$ is the collection of all indices $i$ for which $f_i$ falls into bin $B_k$, and $|I_k|$ denotes the number of elements in $I_k$. The corresponding fraction of observations that fall in the $K^{th}$ bin is given by

$$\bar{F}_{Y_k}(z) = \frac{1}{|I_k|} \sum_{I_k} F_{Y_i}(z), \text{ where } F_{Y_i}(z) = \begin{cases} 1 & Y_i > z; \\ 0, & otherwise. \end{cases} \tag{12}$$

The reliability diagram plots $\bar{F}_{f_k}(z)$ against $\bar{F}_{Y_k}(z)$ for the total number of forecasts in each bin.

## 3. Results and Discussion

We first generate the hydrologic model simulations (Figure 3). Streamflow simulations are obtained by forcing the hydrologic model with gridded observed precipitation and near-surface temperature data. The performance of the hydrologic model simulation is satisfactory, with the NSE of 0.67 and an underestimation bias of -13% (Figure 3). High flows typically demonstrate a larger underestimation bias. This is not surprising since the high flows often result from the direct response of the basin to extreme precipitation events. Uncertainties associated with the precipitation estimates could propagate to the streamflow predictions. High flows predictions can particularly benefit from improving the quality of meteorological forcings (e.g., temperature and precipitation), good representation of hydrologic model states, parameters, and better initial conditions. Standalone LSTM improves streamflow simulations over the hydrologic model simulations. Relative to the standalone LSTM and hydrologic model simulations, LSTM postprocessed simulations improve model performance metrics. LSTM postprocessor improves the NSE and substantially reduces the underestimation bias.

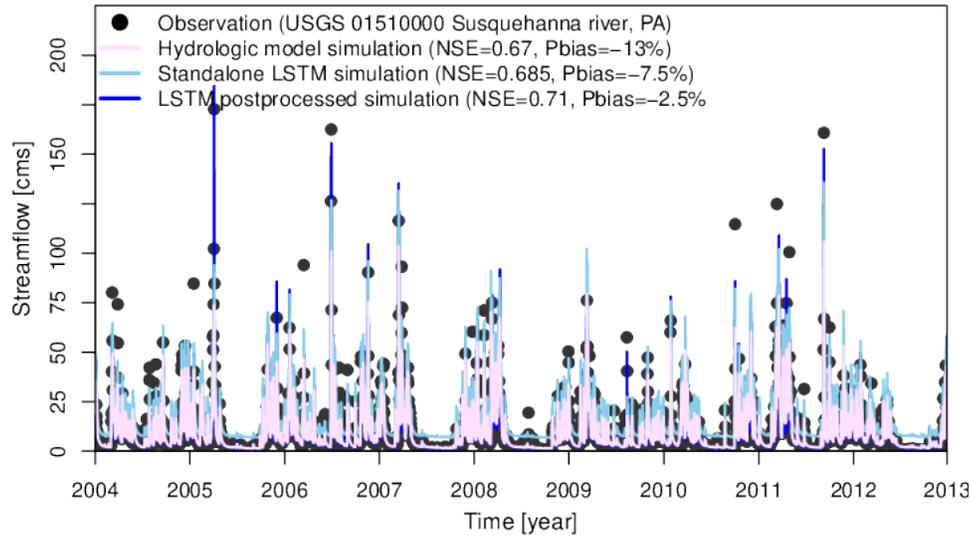

**Figure 3:** Comparison of simulated and observed streamflow hydrographs. Daily streamflow observations are from the US Geological Survey (USGS) at USGS 01510000 located in the Susquehanna River, Pennsylvania, US. We also show the model performance metrics: correlation coefficient (R), Nash-Sutcliffe efficiency coefficient (NSE), and Percent bias (Pbias).

Figure 4 shows the NSE and RMSE of the persistence-based forecast, deterministic forecast, the raw ensemble mean forecast, and postprocessed ensemble mean forecast as a function of forecast lead time. Overall, streamflow forecast quality tends to decline with increasing lead time (Figure 4). Raw ensemble mean streamflow forecasts outperform both the deterministic and persistence-based forecasts. Note, however, that the persistence forecasts are the lower-cost benchmark forecast system, and are often considered a "hard-to-beat" method at medium-range timescales, particularly for larger basin scales (Ghimire and Krajewski 2020; Krajewski et al. 2020; Krajewski et al. 2021). As compared to the deterministic forecasts, improvements from the ensemble mean tend to be higher at longer forecast lead times. This improvement is generally driven by the GEFSRv2 ensemble meteorological forcing since we have not used any ensembles to account for uncertainties issued from hydrologic sources.

The most salient feature in Figure 4 is that LSTM postprocessed ensemble mean forecast outperforms all other forecasting approaches across all the lead times. The improvement is small at the initial lead time but gradually increases with longer forecast lead times. This is expected because LSTM tends to reduce overall uncertainty which shows an increasing trend with increased lead times. At longer lead times, thus, LSTM postprocessed forecast has shown the ability to remove greater amounts of biases when compared to other techniques (Figure 4b). As compared to the raw ensembles generated by hydrometeorological modeling, the relative gain in NSE from the LSTM postprocessor varies from ~0.03 on day 1 to ~0.43 on day 7. The general tendency is for hydrometeorological modeling and standalone LSTM to perform similarly. The standalone LSTM, however, tends to show a slight NSE gain at later forecast lead times. As compared to the standalone LSTM, the gain in NSE from the LSTM postprocessor is as high as 0.38 at the lead time of day 7. A slight jump in the deterministic metrics (NSE and RMSE) for both raw and LSTM-postprocessed ensemble mean forecast at day 4 could be due to the inability of the ensemble mean to capture some important features of forecast uncertainty.

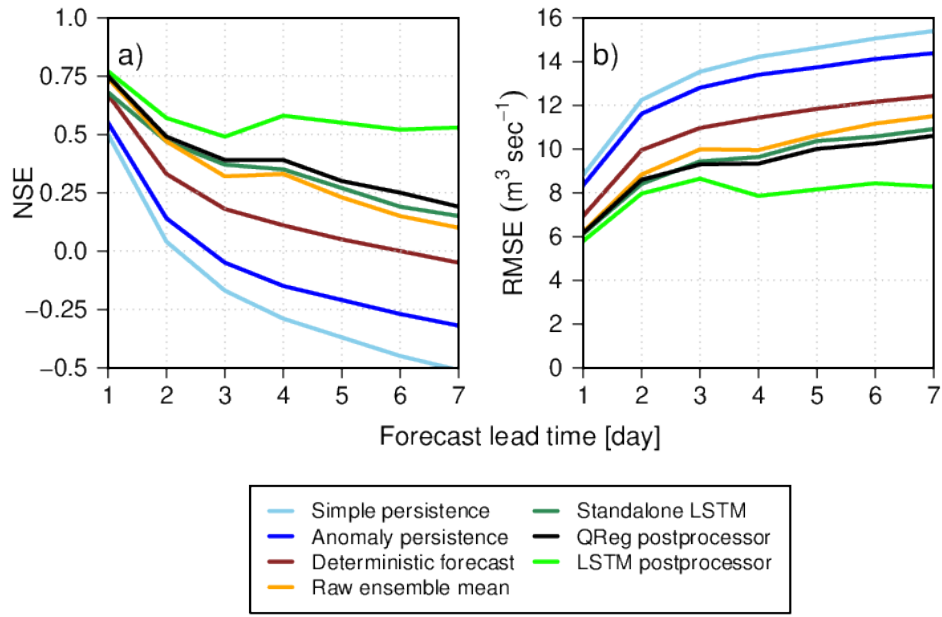

**Figure 4:** a) Nash-Sutcliffe efficiency (NSE) and b) Root mean square error (RMSE) between forecast and corresponding observed streamflow.

Furthermore, we compare the performance of LSTM with the regression-based quantile regression postprocessor (Koenker, 2005). We select the quantile regression postprocessor for comparison since it is the most widely used hydrologic postprocessor (Mendoza et al. 2016; Dogulu et al. 2015; López López et al. 2014; Weerts et al. 2011) and has shown to outperform several postprocessors for different forecasting conditions (Sharma et al. 2019). Quantile regression has several strengths (Koenker, 2005): (i) no prior assumptions regarding the shape of the distribution; (ii) provides conditional quantiles rather than conditional means, and (iii) less sensitive to the tail behavior of the streamflow dataset and consequently, less sensitive to outliers. We used quantile regression to estimate error distribution, which is then added to the ensemble mean to form a calibrated discrete quantile relationship for a particular lead time and generate an ensemble streamflow forecast (Dogulu et al., 2015; López López et al., 2014; Weerts et al., 2011). Overall, the LSTM postprocessor performs better than quantile regression across all the lead times (Figure 4a-b). The differences in NSE and RMSE between the two postprocessors are minimal at initial lead times (days 1 and 2). However, as the lead time progresses, the LSTM shows substantial improvement over quantile regression. As compared to the quantile regression, the relative gain in NSE from LSTM varies from 0.02 on day 1 to as high as 0.34 on day 7 (Figure 4a). Overall, the NSE of LSTM-postprocessed ensemble mean forecasts ranges from ~0.77 (day 1) to ~0.52 (day 7) (Figure 4a); while the RMSE ranges from ~6.0 m$^3$/s (day 1) to ~8.5 m$^3$/s (day 7) (Figure 4b).

Figure 5 examines the skill by estimating the BSS in ensemble streamflow forecast conditioned up on the season, lead time, and flow threshold. The BSS is computed with reference to the climatological forecast. For forecast verification, two seasons: cool (October – March; Figure 5a) and warm (April-September; Figure 5b) have been taken into consideration for lead times from 1 to 7 days. The low-moderate flow category in Figure 5 represents flows with a non-exceedance probability of 0.50, while the high-flow category is for a non-exceedance probability of 0.90 (i.e., flows with exceedance probability less than 0.1 are denoted as high).

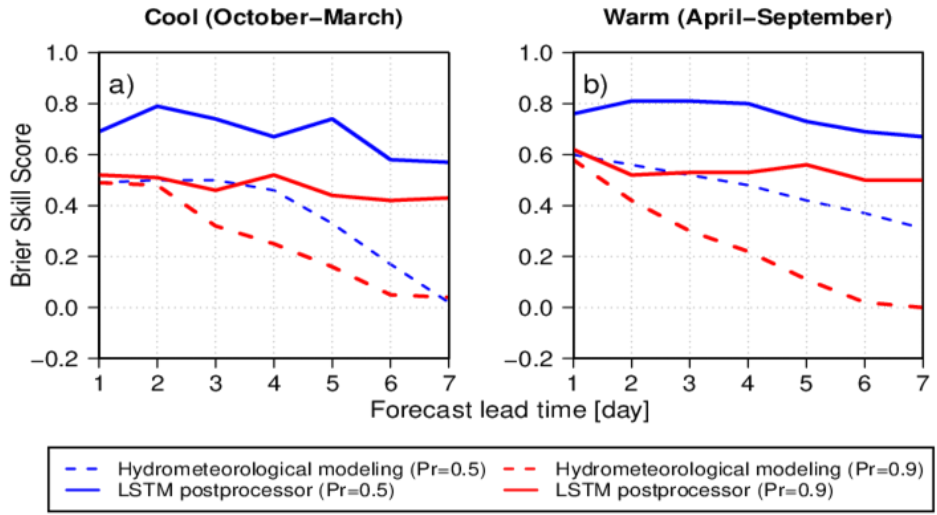

**Figure 5:** Brier Skill Score (BSS) of the raw ensemble forecasts (dashed lines) and LSTM postprocessed ensemble forecasts (solid lines) vs. the forecast lead time during the a) cool (October-March) and b) warm seasons (April-September), under low-moderate flow and high flow conditions. Raw ensemble forecasts are generated using the hydrologic model forced with the precipitation and near-surface temperature ensemble forecasts from the National Centers for Environmental Prediction Global Ensemble Forecast System Reforecast version 2 (GEFSRv2). The low-moderate flow category represents flows with a non-exceedance probability of 0.50 (Pr=0.5), while the high-flow category is for a non-exceedance probability of 0.90 (Pr=0.9; i.e., flows with exceedance probability less than 0.1 are denoted as high).

The skill of the postprocessed ensemble streamflow forecasts is greater than the raw ones across seasons, flow thresholds, and lead times (Figure 5). As expected from our previous results (Figure 4), the relative gain in skill from LSTM increases with forecast lead times. This indicates that the streamflow forecasts are influenced by systematic biases, and those biases appear to have a strong effect at longer lead times. Indeed, as compared to the raw ensemble, skill dependence with lead times is reduced after postprocessing. Seasonal improvements in postprocessed forecast skills to capture baseflow or low flow have been remarkably good across all the lead times. This means LSTM can skillfully predict low to moderate flows irrespective of seasonal influences. This is understandable as hydrologic systems do not show significant alterations during low flow events. However, for the high flows, the skill improvement from LSTM becomes more apparent at later forecast lead times demonstrating its added value in forecasting flood conditions at medium-range time scales. High flows result from the direct response of the basins to extreme precipitation events, whereas the low to moderate flows are dominated by subsurface processes. In addition, the hydrologic uncertainties dominate forecast skill at initial lead times, while the meteorological uncertainties are more influential at longer forecast lead times (Siddique and Mejia 2017). Hence, the high flow forecast at medium-range timescales and smaller basins can particularly benefit from improved ensemble meteorological forcing (Ghimire et al., 2021); whereas the forecast skill at initial lead times can benefit from the improved representation of hydrologic model states (e.g., soil moisture) and initial conditions. This highlights the need of addressing both hydrologic and meteorologic uncertainties to further enhance the streamflow forecast skill. As our results show, LSTM can significantly complement hydrologic models in addressing such a challenge. The results are promising because the LSTM derives its enhanced skill from the streamflow persistence in addition to the long memory of the meteorological forcing that LSTM preserves.

The overall skill of both raw and postprocessed forecasts is slightly greater in the warm season than in the cool ones (Figure 5). Seasonal skill differences are more apparent for raw ensemble streamflow forecasts and particularly for low-moderate flows. The reason for seasonal skill variations is that during the cool season the hydrologic conditions are likely influenced by snow accumulation and melting. Thus, a better representation of snow dynamics in hydrologic modeling could contribute to improving the forecast skill. As compared to the raw ensemble, seasonal skill variations for low-moderate flows are somewhat reduced after postprocessing. Note that LSTM benefits from hydrologic persistence which is generally stronger for low-moderate flow conditions (Ghimire and Krajewski, 2020). Overall, for low-moderate flow conditions, the BSS of the postprocessed forecast ranges from ~0.69 (day 1) to ~0.57 (day 7) during the cool season (Figure 5a), and 0.76 (day 1) to 0.67 (day 7) during the warm season (Figure 5b). BSS for high-flow conditions ranges from ~0.52 (day 1) to ~0.43 (day 7) during the cool season (Figure 5a), and 0.62 (day 1) to 0.50 (day 7) during the warm season (Figure 5b).

Figure 6 shows the reliability diagrams for raw and postprocessed ensemble streamflow forecasts for the lead times of 1, 3, and 7 days. The reliability diagrams are based on low-moderate flow and high-flow conditions. As the slope of the estimated reliability curves for raw forecasts tends to be more than the 1:1 reference line, the streamflow forecasts are underconfident across most of the forecast probabilities. This tendency of underconfidence is more apparent at longer forecast lead times similar to results shown in Figures 4 and 5. On day 7 (Figure 6c), raw ensembles are significantly underconfident; they underpredict the larger forecast probabilities. The LSTM-postprocessed forecasts seem relatively more reliable than the raw forecasts across the lead times and a broad range of streamflow thresholds. The relative improvement in reliability by employing LSTM is greater at longer forecast lead times (for example, day 3 and day 7). Overall, the reliability of postprocessed streamflow forecast is higher for medium-range timescales, particularly for the low-moderate streamflow threshold.

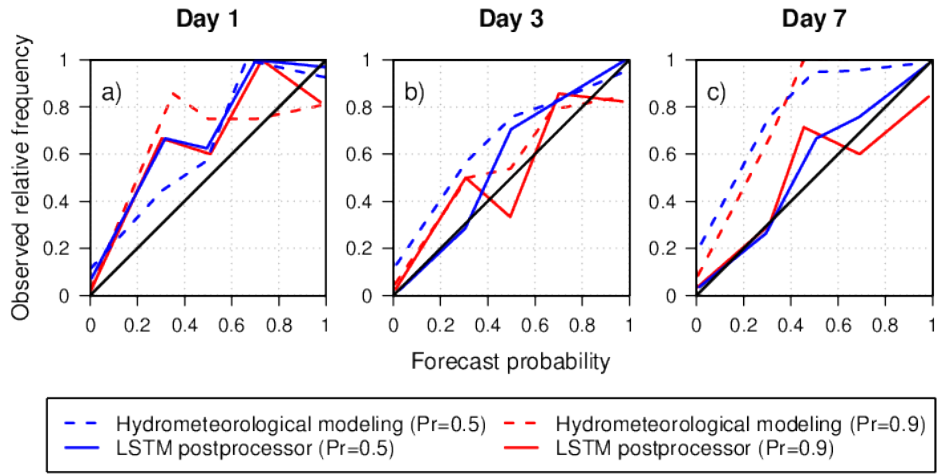

**Figure 6:** Reliability diagrams for the low-moderate flow and high-flow across forecast lead times of a) 1, b) 3, and c) 7 days. Different reliability curves represent the raw and LSTM postprocessed streamflow ensembles. Raw ensemble forecasts are generated using the hydrologic model forced with the precipitation and near-surface temperature ensemble forecasts from the National Centers for Environmental Prediction Global Ensemble Forecast System Reforecast version 2 (GEFSRv2).The low-moderate flow category represents flows with a non-exceedance probability of 0.50, while the high-flow category is for a non-exceedance probability of 0.90 (i.e., flows with an exceedance probability less than 0.1 are denoted as high). Reliability curves that tend to align along the diagonal line are preferred (more reliable).

## 4. Conclusion

We demonstrate a case study for machine learning application in postprocessing ensemble streamflow forecasts in a real-time streamflow forecasting setup. We employed the LSTM to correct residual errors and biases in raw ensemble streamflow forecasts generated from a Regional Hydrological Ensemble Prediction System (Sharma et al., 2019) at medium-range (1-7 days lead times) timescales. We configured LSTM for individual 11-member ensemble streamflow forecasts and for each forecast lead time of days 1 to 7. We used LSTM for the period of 2004-2012, with 6 years (2004-2009) for training the LSTM network and the remaining years (2010-2012) for verification. We assessed and verified the performance of LSTM postprocessed forecasts with different forecasting approaches, including simple persistence, anomaly persistence, climatology, deterministic forecast, raw ensemble forecast, and standalone LSTM.

In summary, based on our analysis and comparison, we found that raw ensemble streamflow forecasts generated using hydrometeorological modeling outperform both the deterministic and persistence-based forecasts. The general tendency is for the standalone LSTM to perform slightly better than the raw forecast from hydrometeorological modeling across lead times. LSTM postprocessing improves streamflow simulations and forecasts as compared to other forecasting approaches. LSTM postprocessing can improve forecast skill and reliability across all the seasons, flow thresholds, and forecast lead times. The relative gain from LSTM is generally higher at medium-range timescales (4~7 days lead times) compared to initial lead time (1~3 days lead times); high flows compared to low-moderate flows, and warm-season compared to cool ones. For the high flows, the skill improvement from LSTM becomes more apparent at later forecast lead times demonstrating its added value in forecasting flood conditions at medium-range time scales. The overall skill of both raw and LSTM postprocessed forecasts is slightly greater in the warm season than in the cool ones. The reliability diagram shows that the LSTM postprocessor can correct biases in the raw ensembles ultimately making the postprocessed ensembles more reliable than the raw ones across lead times and flow thresholds.

This case study demonstrates the potential of LSTM to postprocess ensemble streamflow forecasts in a medium-size Upper Susquehanna River basin in the eastern US. Future studies could explore it under a wider range of forecasting conditions and basin scales across diverse hydroclimatic regions. To continue expanding this research, we plan to explore and evaluate the neural network multimodel ensemble to potentially improve flood forecasts across a range of spatiotemporal scales. Multimodel ensemble forecasting can account for multiple hydrometeorological processes interacting nonlinearly and hence could improve the forecast skill over single-model forecasts. A machine learning technique called super-learning or stacked-ensemble (Laan et al. 2007) is a loss-based supervised learning method that allows combining multiple machine learning algorithms. The goal with such implementations would be to bias correct the raw ensemble streamflow forecast using different machine learning techniques, optimally combine the bias-corrected forecasts to generate the super learner ensembles, and compare super learner with the traditional multimodal forecasting approaches such as Bayesian model averaging.

## Acknowledgments

The authors are grateful to the anonymous reviewers for their reviews and constructive comments. This research received no external funding. All authors contributed to the study design. S.S led the calculations and wrote the initial draft of the manuscript. All authors revised and edited the manuscript. The authors declare no competing financial or nonfinancial interests. Daily streamflow observation data for the selected forecast stations can be obtained from the USGS website (https://waterdata.usgs.gov/nwis/). Precipitation and temperature forecast datasets from the Global Ensemble Forecast System Reforecast Version 2 (GEFSRv2) can be obtained from the NOAA Earth System Research Laboratory website (https://www.esrl.noaa.gov/psd/forecasts/reforecast2/download.html).

## References

Alfieri, L., Bisselink, B., Dottori, F., Naumann, G., de Roo, A., Salamon, P., Wyser, K., & Feyen, L. (2017). Global projections of river flood risk in a warmer world. Earth's Future. https://doi.org/10.1002/2016ef000485

Alizadeh, B., Limon, R.A., Seo, D.-J., Lee, H., & Brown, J. (2020). Multiscale Postprocessor for Ensemble Streamflow Prediction for Short to Long Ranges. J. Hydrometeorol. 21, 265–285. https://doi.org/10.1175/JHM-D-19-0164.1

Alizadeh, B., Bafti, A. G., Kamangir, H., Zhang, Y., Wright, D. B., & Franz, K. J. (2021). A novel attention-based LSTM cell post-processor coupled with bayesian optimization for streamflow prediction. Journal of Hydrology, 601, 126526.

Bai, P., Liu, X., & Xie, J. (2021). Simulating runoff under changing climatic conditions: a comparison of the long short-term memory network with two conceptual hydrologic models. Journal of Hydrology, 592, 125779.

Bengio, Y. (2012). Practical Recommendations for Gradient-Based Training of Deep Architectures, in: Montavon, G., Orr, G.B., Müller, K.-R. (Eds.), Neural Networks: Tricks of the Trade: Second Edition. Springer Berlin Heidelberg, Berlin, Heidelberg, pp. 437–478. https://doi.org/10.1007/978-3-642-35289-8_26

Brier, G.W. (1950). Verification of forecasts expressed in terms of probability. Mon. Weather Rev. 78, 1–

3. https://doi.org/10.1175/1520-0493(1950)078<0001:vofeit>2.0.co;2

Brown, J.D., He, M., Regonda, S., Wu, L., Lee, H., & Seo, D.-J. (2014). Verification of temperature, precipitation, and streamflow forecasts from the NOAA/NWS Hydrologic Ensemble Forecast Service (HEFS): 2. Streamflow verification. J. Hydrol. 519, 2847–2868. https://doi.org/10.1016/j.jhydrol.2014.05.030

Cheng, M., Fang, F., Kinouchi, T., Navon, I. M., & Pain, C. C. (2020). Long lead-time daily and monthly streamflow forecasting using machine learning methods. Journal of Hydrology, 590, 125376. https://doi.org/10.1016/j.jhydrol.2020.1

Cho, K. and Kim, Y., 2022. Improving streamflow prediction in the WRF-Hydro model with LSTM networks. Journal of Hydrology, 605, p.127297.

Cloke, H.L., & Pappenberger, F. (2009). Ensemble flood forecasting: A review. Journal of Hydrology. https://doi.org/10.1016/j.jhydrol.2009.06.005

Demargne, J., Wu, L., Regonda, S.K., Brown, J.D., Lee, H., He, M., Seo, D.-J., Hartman, R., Herr, H.D., Fresch, M., Schaake, J., & Zhu, Y. (2014). The Science of NOAA's Operational Hydrologic Ensemble Forecast Service. Bulletin of the American Meteorological Society. https://doi.org/10.1175/bams-d-12-00081.1

Dogulu, N., López López, P., Solomatine, D.P., Weerts, A.H., & Shrestha, D.L. (2015). Estimation of predictive hydrologic uncertainty using the quantile regression and UNEEC methods and their comparison on contrasting catchments. Hydrology and Earth System Sciences. https://doi.org/10.5194/hess-19-3181-2015

Duan, Q., N. K. Ajami, X. Gao, & S. Sorooshian (2007). Multi-model ensemble hydrologic prediction using Bayesian model averaging, Adv. Water Resour., 30(5), 1371–1386, doi:10.1016/j.advwatres.2006.11.014.

Durkee, J.D., Frye, J.D., Fuhrmann, C.M., Lacke, M.C., Jeong, H.G., & Mote, T.L. (2008). Effects of the North Atlantic Oscillation on precipitation-type frequency and distribution in the eastern United States. Theoretical and Applied Climatology. https://doi.org/10.1007/s00704-007-0345-x

Feng, D., Fang, K., & Shen, C. (2020). Enhancing streamflow forecast and extracting insights using long-short term memory networks with data integration at continental scales. Water Resour. Res.

Frame, J. M., Kratzert, F., Raney, A., II, Rahman, M., Salas, F. R., & Nearing, G. S. (2021). Post-processing the National water model with long short-term memory networks for streamflow predictions and model diagnostics. Journal of the American Water Resources Association, 57(6), 885–905. https://doi.org/10.1111/1752-1688.12964

Ghimire, G.R., & Krajewski, W.F. (2020). Exploring persistence in streamflow forecasting. J. Am. Water Resour. Assoc. 56, 542–550. https://doi.org/10.1111/1752-1688.12821

Ghimire, G. R., Krajewski, W.F., & Quintero, F. (2021). Scale-dependent value of QPF for real-time streamflow forecasting. Journal of Hydrometeorology. https://doi.org/10.1175/JHM-D-20-0297.1

Gitro, C.M., Evans, M.S., & Grumm, R.H. (2014). Two Major Heavy Rain/Flood Events in the Mid-Atlantic: June 2006 and September 2011. Journal of Operational Meteorology 2.

Grönquist, P., Yao, C., Ben-Nun, T., Dryden, N., Dueben, P., Li, S., & Hoefler, T. (2021). Deep learning for post-processing ensemble weather forecasts. Philos. Trans. A Math. Phys. Eng. Sci. 379, 20200092. https://doi.org/10.1098/rsta.2020.0092

Hamill, T.M., Bates, G.T., Whitaker, J.S., Murray, D.R., Fiorino, M., Galarneau, T.J., Zhu, Y., & Lapenta, W. (2013). NOAA's Second-Generation Global Medium-Range Ensemble Reforecast Dataset. Bulletin of the American Meteorological Society. https://doi.org/10.1175/bams-d-12-00014.1

Hapuarachchi, H., Bari, M., Kabir, A., Hasan, M., Woldemeskel, F., Gamage, N., & Feikema, P. (2022). Development of a national 7-day ensemble streamflow forecasting service for Australia. Hydrol. Earth Syst. Sci. Discuss. 2022, volume 2, 1–35.

Hochreiter, S., & Schmidhuber, J. (1997). Long Short-Term Memory. Neural Computation. https://doi.org/10.1162/neco.1997.9.8.1735

Hunt, K. M., Matthews, G. R., Pappenberger, F., & Prudhomme, C. (2022). Using a long short-term memory (LSTM) neural network to boost river streamflow forecasts over the western United States. Hydrology and Earth System Sciences Discussions, 1-30.

Jolliffe, I.T., & Stephenson, D.B. (2012). Forecast Verification: A Practitioner's Guide in Atmospheric Science. John Wiley & Sons.

Kingma, D.P., & Ba, J. (2014). Adam: A Method for Stochastic Optimization. arXiv [cs.LG].

Kim, S., Sadeghi, H., Limon, R. A., Saharia, M., Seo, D.-J., Philpott, A., Bell, F., Brown, J., & He, M. (2018). Assessing the skill of medium-range ensemble precipitation and Streamflow forecasts from the Hydrologic Ensemble Forecast Service (HEFS) for the upper Trinity River basin in North Texas. Journal of Hydrometeorology, 19(9), 1467–1483. https://doi.org/10.1175/JHM-D-18-0027.1

Kirkwood, C., Economou, T., Odbert, H., & Pugeault, N. (2021). A framework for probabilistic weather forecast post-processing across models and lead times using machine learning. Philos. Trans. A Math. Phys. Eng. Sci. 379, 20200099. https://doi.org/10.1098/rsta.2020.0099

Koenker, R. (2005). Quantile Regression. https://doi.org/10.1017/cbo9780511754098

Konapala, G., Kao, S.-C., Painter, S.L., & Lu, D. (2020). Machine learning assisted hybrid models can improve streamflow simulation in diverse catchments across the conterminous US. Environ. Res. Lett. 15, 104022. https://doi.org/10.1088/1748-9326/aba927

Koren, V., Reed, S., Smith, M., Zhang, Z., & Seo, D.-J. (2004). Hydrology laboratory research modeling

system (HL-RMS) of the US national weather service. J. Hydrol. 291, 297–318. https://doi.org/10.1016/j.jhydrol.2003.12.039

Krajewski, W.F., Ghimire, G.R., & Quintero, F. (2020). Streamflow Forecasting without Models. J. Hydrometeorol. 21, 1689–1704. https://doi.org/10.1175/JHM-D-19-0292.1

Krajewski, W.F., Ghimire, G. R., Demir, I., & Mantilla, R. (2021). Real-time streamflow forecasting: AI vs. Hydrologic insights. Journal of Hydrology X. https://doi.org/10.1016/j.hydroa.2021.100110

Kratzert, F., Klotz, D., Brenner, C., Schulz, K., & Herrnegger, M. (2018). Rainfall–runoff modelling using Long Short-Term Memory (LSTM) networks. Hydrol. Earth Syst. Sci. 22, 6005–6022. https://doi.org/10.5194/hess-22-6005-2018

Kratzert, F., Klotz, D., Herrnegger, M., Sampson, A.K., Hochreiter, S., & Nearing, G.S. (2019). Toward improved predictions in ungauged basins: Exploiting the power of machine learning. Water Resour. Res. 55, 11344–11354. https://doi.org/10.1029/2019wr026065

Kumar D., Singh A., Samui P. and Jha R. K. (2019). Forecasting monthly precipitation using sequential modelling Hydrol. Sci. J. 64 690–700

Laan, M.J. van der, van der Laan, M.J., Polley, E.C., & Hubbard, A.E. (2007). Super Learner. Statistical Applications in Genetics and Molecular Biology. https://doi.org/10.2202/1544-6115.1309

Le, Xuan-Hien, Hung Viet Ho, Giha Lee, and Sungho Jung. "Application of long short-term memory (LSTM) neural network for flood forecasting." Water 11, no. 7 (2019): 1387.

Lee, D. G., & Ahn, K. H. (2021). A stacking ensemble model for hydrological post-processing to improve streamflow forecasts at medium-range timescales over South Korea. Journal of Hydrology, 600, 126681.

Liu, J., Yuan, X., Zeng, J., Jiao, Y., Li, Y., Zhong, L., & Yao, L. (2022). Ensemble streamflow forecasting over a cascade reservoir catchment with integrated hydrometeorological modeling and machine learning. Hydrology and Earth System Sciences, 26(2), 265-278.

Loken, E.D., Clark, A.J., & Karstens, C.D. (2020). Generating Probabilistic Next-Day Severe Weather Forecasts from Convection-Allowing Ensembles Using Random Forests. Weather and Forecasting. https://doi.org/10.1175/waf-d-19-0258.1

Loken, E.D., Clark, A.J., McGovern, A., Flora, M., & Knopfmeier, K. (2019). Postprocessing Next-Day Ensemble Probabilistic Precipitation Forecasts Using Random Forests. Weather Forecast. 34, 2017–2044. https://doi.org/10.1175/WAF-D-19-0109.1

López López, P., Verkade, J.S., Weerts, A.H., & Solomatine, D.P. (2014). Alternative configurations of Quantile Regression for estimating predictive uncertainty in water level forecasts for the Upper Severn River: a comparison. Hydrol. Earth Syst. Sci. Discuss. 11, 3811–3855. https://doi.org/10.5194/hessd-11-3811-2014

Mendoza, P.A., Wood, A., Clark, E., Nijssen, B., Clark, M.P., Ramos, M.H., & Voisin, N. (2016). Improving medium-range ensemble streamflow forecasts through statistical post-processing. 2016 Fall Meeting, San Francisco, CA, Amer. Geophys. Union, Abstract H51F-1547.

Murphy, A. H. (1973). A New Vector Partition of the Probability Score. J. Appl. Meteorol. Climatol. 12, 595–600. https://doi.org/10.1175/1520-0450(1973)012<0595:ANVPOT>2.0.CO;2

Nash, J.E., Sutcliffe, J.V. (1970). River flow forecasting through conceptual models part I — A discussion of principles. Journal of Hydrology. https://doi.org/10.1016/0022-1694(70)90255-6

NWS. (2022). National Weather Service. What is HEFS? https://www.weather.gov/abrfc/about_HEFS

Nearing, G.S., Kratzert, F., Sampson, A.K., Pelissier, C.S., Klotz, D., Frame, J.M., Prieto, C., & Gupta, H.V. (2021). What role does hydrological science play in the age of machine learning? Water Resour. Res. 57. https://doi.org/10.1029/2020wr028091

Pagano, T. C., Elliott, J., Anderson, B., & Perkins, J. (2016). Australian Bureau of Meteorology Flood Forecasting and Warning in: Flood Forecasting, Elsevier, 3–40, 2016.

Panahi, J., Mastouri, R., & Shabanlou, S. (2022). Insights into enhanced machine learning techniques for surface water quantity and quality prediction based on data pre-processing algorithms. Journal of Hydroinformatics.

Rahimzad, Maryam, Alireza Moghaddam Nia, Hosam Zolfonoon, Jaber Soltani, Ali Danandeh Mehr, and Hyun-Han Kwon. "Performance comparison of an lstm-based deep learning model versus conventional machine learning algorithms for streamflow forecasting." Water Resources Management 35, no. 12 (2021): 4167-4187.

Rasp, S., & Lerch, S. (2018). Neural Networks for Postprocessing Ensemble Weather Forecasts. Mon. Weather Rev. 146, 3885–3900. https://doi.org/10.1175/MWR-D-18-0187.1

Regonda, S.K., Seo, D.-J., Lawrence, B., Brown, J.D., & Demargne, J. (2013). Short-term ensemble streamflow forecasting using operationally-produced single-valued streamflow forecasts – A Hydrologic Model Output Statistics (HMOS) approach. Journal of Hydrology. https://doi.org/10.1016/j.jhydrol.2013.05.028

Seo, D.-J., H. D. Herr, & J. C. Schaake (2006). A statistical post-processor for accounting of hydrologic uncertainty in short-range ensemble streamflow prediction, Hydrol. Earth Syst. Sci. Discuss., 3, 1987–2035.

Sharma, S., Siddique, R., Reed, S., Ahnert, P., & Mejia, A. (2019). Hydrological Model Diversity Enhances Streamflow Forecast Skill at Short- to Medium-Range Timescales. Water Resources Research. https://doi.org/10.1029/2018wr023197

Shen, C. (2018). A transdisciplinary review of deep learning research and its relevance for water resources scientists. Water Resour. Res. 54, 8558–8593. https://doi.org/10.1029/2018wr022643

Siddique, R., & Mejia, A. (2017). Ensemble Streamflow Forecasting across the U.S. Mid-Atlantic Region with a Distributed Hydrological Model Forced by GEFS Reforecasts. J. Hydrometeorol. 18, 1905–1928. https://doi.org/10.1175/JHM-D-16-0243.1

Sikorska-Senoner, A. E., & Quilty, J. M. (2021). A novel ensemble-based conceptual-data-driven approach for improved streamflow simulations. Environmental Modelling and Software, 143, 105094. https://doi.org/10.1016/j.envsoft.2021.1

Smith, J.A., Baeck, M.L., Villarini, G., & Krajewski, W.F. (2010). The Hydrology and Hydrometeorology of Flooding in the Delaware River Basin. J. Hydrometeorol. 11, 841–859. https://doi.org/10.1175/2010JHM1236.1

Troin, M., Arsenault, R., Wood, A. W., Brissette, F., & Martel, J. L. (2021). Generating ensemble streamflow forecasts: A review of methods and approaches over the past 40 years, Water Resour. Res., e2020WR028392, https://doi.org/10.1029/2020WR028392.

Tyralis, H., Papacharalampous, G., Burnetas, A., & Langousis, A. (2019). Hydrological post-processing using stacked generalization of quantile regression algorithms: Large-scale application over CONUS. Journal of Hydrology, 577, 123957.

Van, S. P., Le, H. M., Thanh, D. V., Dang, T. D., Loc, H. H., & Anh, D. T. (2020). Deep learning convolutional neural network in rainfall–runoff modelling. Journal of Hydroinformatics, 22(3), 541-561.

Xiang, Z., & Demir, I. (2020). Distributed long-term hourly streamflow predictions using deep learning–A case study for State of Iowa. Environmental Modelling & Software, 131, 104761.

Weerts, A.H., Winsemius, H.C., & Verkade, J.S. (2011). Estimation of predictive hydrological uncertainty using quantile regression: examples from the National Flood Forecasting System (England and Wales). Hydrology and Earth System Sciences. https://doi.org/10.5194/hess-15-255-2011

Wilks, D.S. (2011). Statistical Methods in the Atmospheric Sciences. Academic Press.

Wunsch, A., Liesch, T., & Broda, S. (2021). Groundwater Level Forecasting with Artificial Neural Networks: A Comparison of Long Short-Term Memory (LSTM), Convolutional Neural Networks (CNNs), and Non-Linear Autoregressive Networks with Exogenous Input (NARX), Hydrol. Earth Syst. Sci., 25, 1671–1687, https://doi.org/10.5194/hess-25-1671-2021.

Zhang, J., Chen, J., Li, X., Chen, H., Xie, P., & Li, W. (2020). Combining postprocessed ensemble weather forecasts and multiple hydrological models for ensemble streamflow predictions. Journal of Hydrologic Engineering, 25(1), 04019060.

Zhang D, Lin J, Peng Q, Wang D, Yang T, Sorooshian S, Liu X and Zhuang J. (2018). Modeling and simulating of reservoir operation using the artificial neural network, support vector regression, deep learning algorithm J. Hydrol. 565 720–36

Zhao, L., Q. Duan, J. Schaake, A. Ye, & J. Xia (2011). A hydrologic post-processor for ensemble streamflow predictions, Adv. Geosci., 29, 51–59.